\documentclass{article}

\usepackage{microtype}
\usepackage{graphicx}
\usepackage{subfigure}
\usepackage{booktabs} % for professional tables
\usepackage[ruled,noend]{algorithm2e}
\usepackage{moreverb}
\usepackage{textcomp}
\usepackage{amsmath}
\usepackage{mathtools}
\usepackage{dsfont}
\usepackage{enumitem}
\usepackage{amssymb}
\usepackage{xcolor}
\usepackage{wrapfig}
\usepackage{tabularx}
\newcommand{\somi}[1]{\textcolor{red}{somi: #1}}
\newcommand{\hs}[1]{\textcolor{blue}{hs: #1}}
\newcommand{\E}[2]{\mathbb{E}_{#1}{\left[#2\right]}}

\usepackage{corl_2020} % Use this for the initial submission.
\DeclareMathOperator*{\argmax}{arg\,max} % argmax

\author{
  Shubhankar Agarwal \\
  Uber Advanced Technologies Group\\
  \texttt{sagarwal@uber.com} \\
  \And
  Harshit Sikchi \\
  Carnegie Mellon University,\\
  Uber Advanced Technologies Group\\
  \texttt{hsikchi@uber.com} \\
  \And
  Cole Gulino \\
  Uber Advanced Technologies Group\\
  \texttt{cgulino@uber.com} \\
  \And
  Eric Wilkinson \\
  Uber Advanced Technologies Group\\
  \texttt{ewilkinson@uber.com} \\
}

\begin{document}
\title{Imitative Planning using Conditional Normalizing Flow}
\maketitle{}

%===============================================================================

\begin{abstract}
A popular way to plan trajectories in dynamic urban scenarios for Autonomous Vehicles is to rely on explicitly specified and hand crafted cost functions, coupled with random sampling in the trajectory space to find the minimum cost trajectory. Such methods require a high number of samples to find a low-cost trajectory and might end up with a highly suboptimal trajectory given the planning time budget. We explore the application of normalizing flows for improving the performance of trajectory planning for autonomous vehicles (AVs). Our key insight is to learn a sampling policy in a low-dimensional latent space of expert-like trajectories, out of which the best sample is selected for execution. By modeling the trajectory planner's cost manifold as an energy function we learn a scene conditioned mapping from the prior to a Boltzmann distribution over the AV control space. Finally, we demonstrate the effectiveness of our approach on real world datasets over IL and hand constructed trajectory sampling techniques.
\end{abstract}


%===============================================================================

\section{Introduction}
	
Generating a control trajectory which provides safe, comfortable, and socially responsible motion is a fundamental problem for operating autonomous vehicles (AVs). Since high quality human driving data is easily available, imitative models which learn to mimic expert demonstrations are a popular approach \cite{bansal2018chauffeurnet}. End-to-end imitation learning (IL) approaches are attractive because they allow for a mapping to be learned between high dimensional context features, such as sensor and map data, and the control space of the vehicle platform. 

However, these IL approaches have several limitations which make their use in practice difficult. The first is that for every scene there is only one label, since the expert only provided one demonstration, and it is unclear how to properly penalize deviations from the demonstration. This is the commonly known distribution shift problem~\cite{ross2011reduction}, and a lack of an accurate simulator precludes us from correcting the distribution shift. The second is that the internal belief state of the expert is not available which means the AV is unlikely to learn the correct response to it's own aleatoric and epistemic uncertainties about a road scene. Finally, AV operation typically requires high confidence in the safety outcomes of a control trajectory, which typically necessitates a downstream whitebox costing module to certify the IL method's output. 

In this work, we propose a method to address these problem by treating the whitebox costing module as an energy based model and learning a sampling policy that minimizes certain $f$-divergence to it. Furthermore, we restrict the policy actions to a lower dimensional latent space, which is trained to encode trajectories obtained from the expert demonstrations. Whitebox planners ingest interpretable representations of the scene, which enables the enforcement of strong conditions on safety, and can reason about the uncertainties of the AV system. Additionally, in contrast to a single expert demonstration, the cost manifold provides information about how to penalize deviations from the optima. Thus instead of learning the PDF of the expert given a scene the proposed method learns a density function which corresponds to the planner's cost manifold. 
 
% A trajectory planner uses a cost manifold over the control trajectory space to define the desired AV behavior as the global optima of the manifold.

Our approach builds upon normalizing flows which are capable of representing complex, multimodal manifolds from a known prior distribution and supports efficient, parallel sampling. First, we use a variational autoencoder to learn a representative subspace of the control trajectory space from all expert driving demonstrations. Samples from this encoding space should generate control trajectories which behave \emph{stylistically} the same as the expert, or encode trajectories that are kinematically similar to expert. Then we learn a normalizing flow mapping from the prior distribution to a Boltzmann distribution in the control trajectory encoding space using the cost manifold as an energy function. We propose using neural autoregressive flow (NAF) \cite{huang2018neuralaf} for this flow mapping because of it's ability to learn more complicated multimodal target distribution, while performing  accurate PDF inference. We train our flow method as in inverse autoregressive flow (IAF)  \cite{kingma2016improved} which allows for efficient control trajectory sample generation using parallel transformations. We refer to our method subsequently as \textit{FlowPlan}.

The main contributions of this work are
\somi{
\begin{itemize}[leftmargin=*]
\itemsep0em
\item An efficient method to generate trajectories for Autonomous driving by learning a sampling policy in a scene-conditioned low-dimensional latent space representative of expert driving demonstrations.
% \item A method to construct a scene-conditioned low-dimensional latent space , and use it for efficient learning of a sampling policy.
\item We demonstrate the utility of normalizing flow by taking advantage of the accurate pdf inference to further refine our generated trajectories without the whitebox costing module.
\item As a by-product of our sampling policy we can efficiently generate scores (log probs) of the sampled trajectories without the whitebox costing module.
\item We demonstrate the benefits of our approach over hand constructed, parametric sampling strategies on real world datasets.
\end{itemize}}

%===============================================================================

\section{Related Work}
\label{sec:related_work}

\textbf{Trajectory sampling} techniques for planning attempt to construct trajectories from structured, parametric representations which are likely to solve the SDV's planning problem. One common method used for in-lane driving is to construct samples within a Frenet frame around a nominal path as explored by  \cite{werling2010optimal} with traffic-adaptive velocity profiles for highway driving. A review encompassing these approaches including clothoid, bezier, and polynomial representations can be found in  \cite{gonzalez2015review}. In contrast to our approach, these methods typically involve hand crafting strategies for adapting the parameters of the trajectory representation to the planning problem.  

\textbf{Variational methods} which perform continuous optimization in a function space are typically solved with iterative strategies such as DDP \cite{jacobson1970differential} or iLQR  \cite{li2004iterative}. These methods can only represent a small subset of the real world problems, i.e. convex problems or quadratic in case of iLQR, while most of the self-driving problems are non-convex. A survey of this class of approaches can be found in \cite{betts1998survey}. Our work complements these methods since it can target multimodal (non-convex) cost surfaces to efficiently find locally optimal solutions and can be used to warm-start for optimization algorithms for faster convergence. 

\textbf{Imitation Learning Methods}
Learning-based approaches have also been used for motion planning, Imitation Learning (IL) being one of the most popular ones of those. Expert demonstrations are used to learn the desired behavior or driving policy. \cite{Pomerleau1988ALVINNAA} being one of the first successful implementations. Since then, significant progress has been made to accomplish more complex maneuvers and scenarios, in \cite{codevilla2018endtoend, liang2018cirl}. But, these approaches are not able to generalize outside the expert demonstrations as shown in \cite{Codevilla2019limitations}. \cite{bansal2018chauffeurnet} and \cite{Tigas2019RobustIP}, address generalization outside expert demonstrations by doing closed-loop training and adding different goal functions to guide imitation policy respectively. While these IL approaches alleviate the need for hand-tuning cost functions, they suffer from compounding errors due to auto-regressive nature and provide very little or no interpretability. There has also been parallel work, which explores normalizing flows for learning the density function. \cite{Bhattacharyya2019ConditionalFV} combine a conditional normalizing flow model with VAE to learn an invertible density model for trajectory sampling from expert demonstration. \cite{Mangalam2020endpoint} investigate conditional VAEs with end-point conditioning to accomplish goal-directed sampling along with a social pooling layer for capturing interaction.
\somi{Mention how we are different.}

\textbf{Inverse Reinforcement Learning}
Inverse Reinforcement Learning (IRL) based approaches have been used to learn the motion planning cost functions (reward functions), removing the need for hand constructing cost surfaces to drive behavior.  Since being proposed for the first time for autonomous driving~\cite{abbeel2004irl}, IRL has been used for motion planning in \cite{sun2018hirl, you2019rlirl} and \cite{levine2012ioc}. Most of these approaches use IRL to make discrete decisions (pass, yield, etc.) and operate in very specific simulated scenarios with significantly smaller feature space than the real world. IRL methods like  ~\cite{ziebart2008maximum,fu2017learning} require access to a simulator for extracting the cost functions. Since it is infeasible to let the AV explore in the real world, the use of a simulator is required, and simulator inaccuracy can further lead to sim-to-real transfer issues.

%===============================================================================

\section{Background}\label{sec:background}

\subsection{Motion Planning Problem}\label{section:planning_problem}
The purpose of the planner is to provide safe, comfortable motion for an autonomous vehicle constrained by dynamic and kinematic feasibility, partial observability, and user experience preferences. This is accomplished by formulating the problem as a partially observable Markov decision process (POMDP) which is optimized over a finite time horizon $T$. In this work the POMDP model is defined by the tuple $\langle \mathcal{S, A, O, T, Z, C}, b \rangle$ where $\mathcal{S}$ is the state space of the scene, $\mathcal{A}$ is the action space, $\mathcal{O}$ is the observation space, and $\mathcal{T}(s' | s, a)$ is the probabilistic transition function  from state $s$ to $s'$ when taking action $a$. \hs{What is Z and C?}. The belief state $b_t(s)$ is a probability distribution over the scene states $s \in \mathcal{S}$ which the AV maintains from the history of observations and actions $h_t = (o_0, a_0, o_1, a_1,  \ldots, o_{t-1}, a_{t-1})$ and the initial belief state $b_0$. The observation and transition models allow for the belief state to be updated through Bayes rule. A complete description of POMDPs can be found in \cite{kochenderfer2015decision}.   
% \hs{The planning objective is typically formulated as online trajectory optimization:
% \begin{equation}\label{eq:planning_objective}
%     \tau^* = \text{argmin}_{\tau}~\E{\mathcal{T}(s'|s,a)}{(\sum_{t=1}^{T-1}\mathcal{C}(s_t, a_t | b)) + \mathcal{C}_T(s_T, a_T | b)]}
% \end{equation}}

\hs{In this work the policy $\mathbf{\pi}$ is a stochastic mapping 
$\mathcal{B} \rightarrow \mathcal{A}^{T}$ from belief space to the action sequence of horizon T. We formulate the planner cost as an energy based model \cite{lecun2006tutorial} which define a Boltzmann distribution using exponentiated cost functions i.e $\pi(\textbf{a}|b) \propto  \prod_{t=0}^{T-2} e^{-\mathcal{C}(s_t, a_t | \mathbf{\pi}, b_t)} \cdot e^{-\mathcal{C}_T(s_{T-1}, a_{T-1} | \mathbf{\pi}, b_T)}$ where $\textbf{a}$ is the action sequence and $\mathcal{C}_T$ is a terminal cost function that approximates the remaining cost-to-go. The performance of the policy is given by:}

\begin{equation}\label{eq:energy_function}
    J(\mathbf{\pi} | b) = \E{s_0\sim b, s_t\sim \mathcal{T}, \textbf{a}\sim\pi(.|b)}{\frac{1}{Z}\prod_{t=0}^{T-2} e^{-\mathcal{C}(s_t, a_t | \mathbf{\pi}, b_t)} \cdot e^{-\mathcal{C}_T(s_{T-1}, a_{T-1} | \mathbf{\pi}, b_T)} }
\end{equation}
\hs{The planner performs an online search for the optimal deterministic policy $\mathbf{\pi}^*$ which maximizes the expected value of the distribution under the belief state $b$}
\begin{equation}   \label{eq:optimal_pi_finite_horizon}
    \mathbf{\pi}^{*}= \underset{\mathbf{\pi} \in \mathcal{P}}\argmax~ J(\mathbf{\pi}| b) 
\end{equation}

% In this work the policy $\mathbf{\pi}$ is a deterministic mapping 
% $\mathcal{B} \rightarrow \mathcal{A}$ from belief space to the action space. We formulate the planner cost as an energy based model \citep{lecun2006tutorial} which define a Boltzmann distribution from the product of exponentiated cost functions

% \begin{equation}\label{eq:energy_function}
%     J(\mathbf{\pi} | b) = \frac{1}{Z}\prod_{t=0}^{T-1} e^{-\mathcal{C}(s_t, a_t | \mathbf{\pi}, b_t)} \cdot e^{-\mathcal{C}_T(s_T, a_T | \mathbf{\pi}, b_T)} 
% \end{equation}

% \noindent
% Where $\mathcal{C}_T$ is a terminal cost function that approximates the remaining cost-to-go. The planner performs an online search for the optimal policy $\mathbf{\pi}^*$ which maximizes the expected value of the distribution under the belief state $b$

% \begin{equation}   \label{eq:optimal_pi_finite_horizon}
%     \mathbf{\pi}^{*}= \underset{\mathbf{\pi} \in \mathcal{P}}\argmax~\E{\mathcal{T}}{ J(\mathbf{\pi}| b) }
% \end{equation}

\noindent
% Candidate policies $\mathbf{\pi}$ must reside in the set of dynamically feasible policies $\mathcal{P}$. The obtained solution can be interpreted as a maximum entropy sampling distribution for the whitebox planner. A maximum entropy policy can be proved to be a solution of robust-reward control problem in the presence of an adversary as proved in \citet{eysenbach2019if}. Even in the setting without an adversary, the adversarial objective bounds the worst case performance of the agent. 

% \begin{align*}
%     J(s | \mathbf{\pi}, b(s)) & = \sum_{t=0}^{T-1} C(s_t, a_t | \mathbf{\pi}, b) + C_T(s_T, a_T | \mathbf{\pi}, b)                                                  \\
%     \mathbf{\pi}^{*}          & = \underset{\mathbf{\pi} \in \mathcal{P}}\argmin \EX[ J(s | \mathbf{\pi}, b(s)) ] \numberthis  \label{eq:optimal_pi_finite_horizon}
% \end{align*}

\subsection{Normalizing Flows}\label{section:normalizing_flows}

% \begin{figure}[t]
%   \centering
%   \includegraphics[width=1.0\linewidth]{figures/normflow1.png}
%   \caption{\textbf{Top row} shows transformations of distributions $q$ and $p_Y$ as $g$ and $f$ are applied. \textbf{Bottom Left} shows the invertible transformation $f$. \textbf{Bottom Right} shows the absolute Jacobian of $f$. Fig from  \citep{kobyzev2019nfreview}}.
%   \label{fig:flow_fig1}
% \end{figure}

A finite normalizing flow (flow) is an iterative framework for estimating and building flexible target distributions introduced in \cite{rezede2015nf}. The flow model consists of a series of invertible transformations $\tau_n$ which map a known prior distribution $q(z_0)$ to a potentially complex, target distribution while preserving the total probability mass of the original $pdf$.
More formally,

\begin{equation}\label{eq:flow_prior_sample}
    z_0 \sim q(z_0)
\end{equation}
\begin{equation}\label{eq:flow_transforms}
        z_N = \tau_n(z_{n-1}; \theta | h_t) , \quad \forall n = 1....N
\end{equation}

where $\theta$ are the parameters of the flow model transformations and depend \somi{conditioned} upon action and observation history $h_t$. Since each transformation is invertible, we can use the change of variables formula to obtain the final log density: 
\begin{equation}\label{eq:flow_log_density}
    \log q(z_N|h_t) = \log q(z_0|h_t) - \sum_{n=1}^{N} \log \det |\frac{dz_n}{dz_{n-1}} | 
\end{equation}

We can think of transformations $\tau_n$
as expanding or contracting the space of the known prior $q(z_0)$ into the conditional target $q(z_N | h_t)$ with the corresponding Jacobian determinant describing the relative change of volume and ensuring total probability mass is conserved.

%===============================================================================

\begin{figure*}
  \centering
  \includegraphics[width=1\linewidth,scale=5]{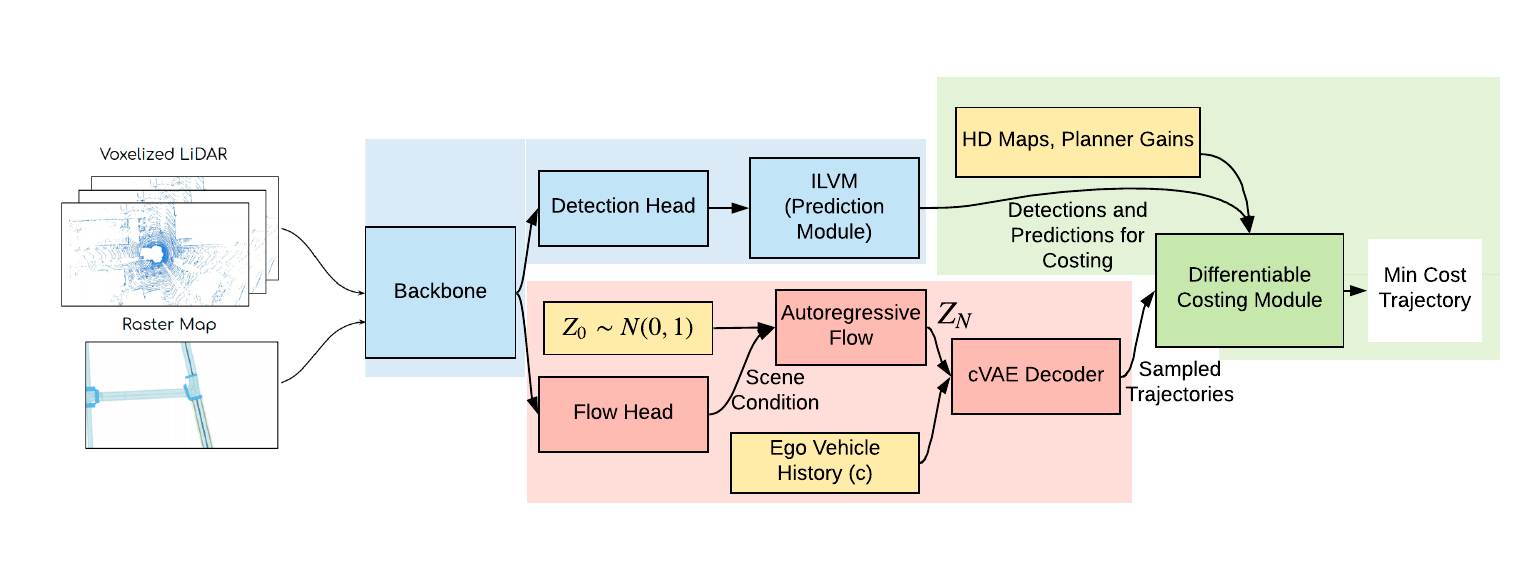}
  \caption{Our model architecture for FlowPlan. Blue modules are used from previous work~\citep{yang2018pixor, casas2020implicit} which are pretrained and are kept frozen during training. Red modules represent our flow planner which works as a control trajectory sampling module. Green modules represent the components of a traditional trajectory planner.}
  \label{fig:architecture}
\end{figure*}

\section{Method}

An overview of our model architecture for FlowPlan can be found in Figure \ref{fig:architecture}. Raw sensor data (LiDAR, Cameras, Radars) and HD map data is processed by a backbone network, to construct an internal feature representation. Actor detections and future predictions are generated from the output of backbone network using separate deep networks, described in section \ref{section:scene_conditioning}. \hs{The detections, predictions and HD maps are used by the whitebox costing module to provide the cost for each trajectory and is described in section \ref{section:traj_planner}.} A $\sigma$VAE is used to learn a reduced dimensional latent space of the trajectory control samples from expert \hs{human} demonstrations, described in section \ref{section:vae}. Our flow network works in parallel to the detector head and also consumes the output of backbone network as a scene conditioner. An autoregressive flow conditioned on the scene, generates trajectory samples in a latent space, described in section \ref{section:sampling_policy}. The flow module is trained to minimize the loss function defined in section \ref{section:sampling_policy}.    

% We begin by formalizing our assumptions and notations. 
% Ego vehicle state $s$ and controls $a$ at time $t$ is $s_t, a_t \in \mathds{R}^{D} $. $s_T$ and $a_T$ represent states and controls for time horizon $T$ secs, similarly, $s_{-L}$ and $u_{-L}$ represent states and controls $L$ secs in history. 
% $z$ represents the latent variable used in cVAE latent space and autoregressive flow. $z_n$ represents the $n^{th}$ step in autoregressive flow. $\overline{s} \in \mathds{R}^{D} $ represents the scene condition as described in section \ref{section:scene_conditioning}.

\subsection{Scene Conditioning}\label{section:scene_conditioning}
% \begin{wrapfigure}{r}{0.4\textwidth}
%   \centering
%   \includegraphics[width=0.96\linewidth]{figures/beliefs/belief1.png}
%   \caption{Interpretable belief representation form the context vector which is input to the flow network and used for sampling trajectories.}
%   \label{fig:CEM}
% \end{wrapfigure}

\begin{figure*}
\vspace{1mm}
    \centering
    \begin{tabular}{cc}
    \includegraphics[width=0.48\textwidth]{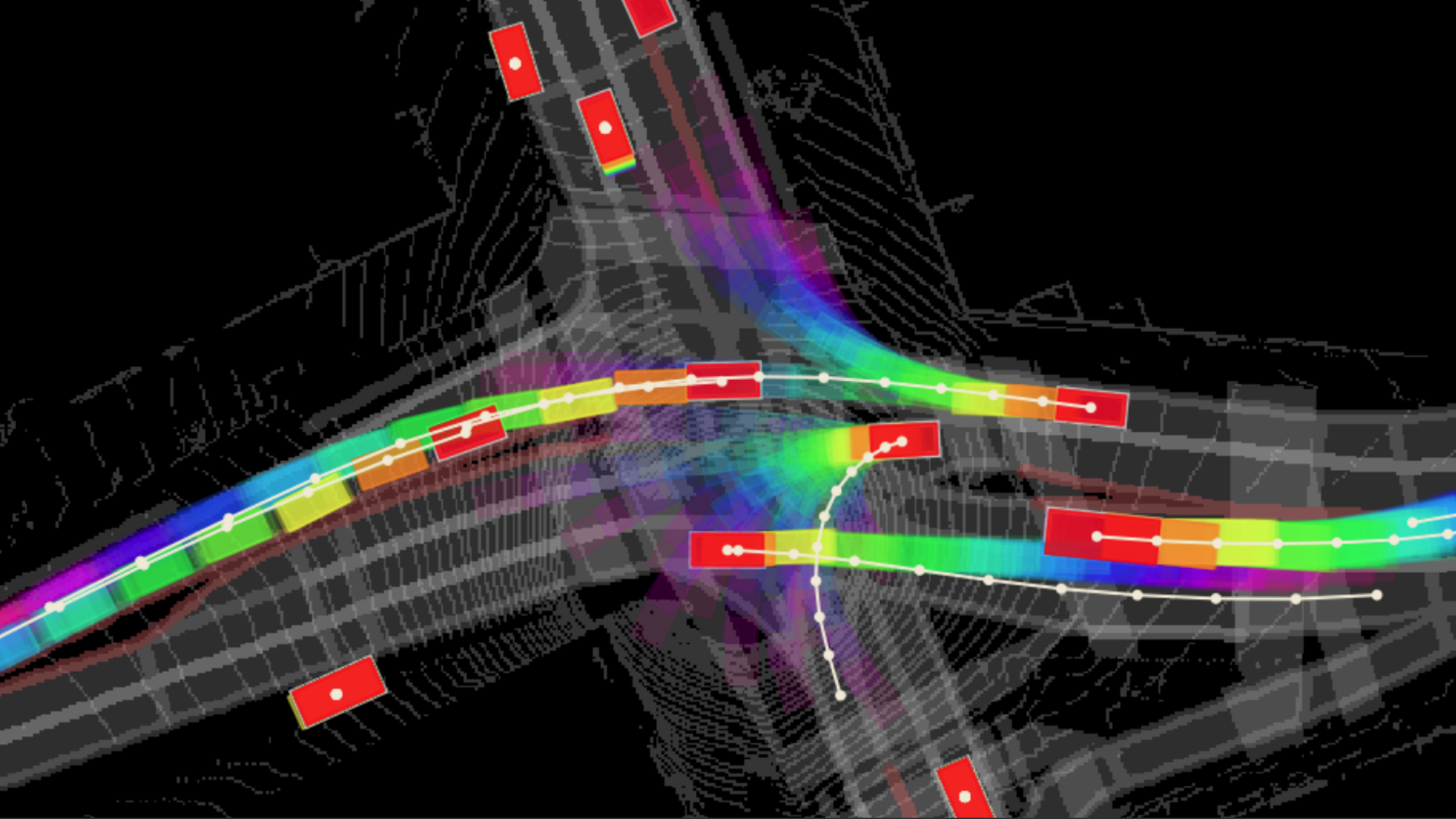}
    & \includegraphics[width=0.48\textwidth]{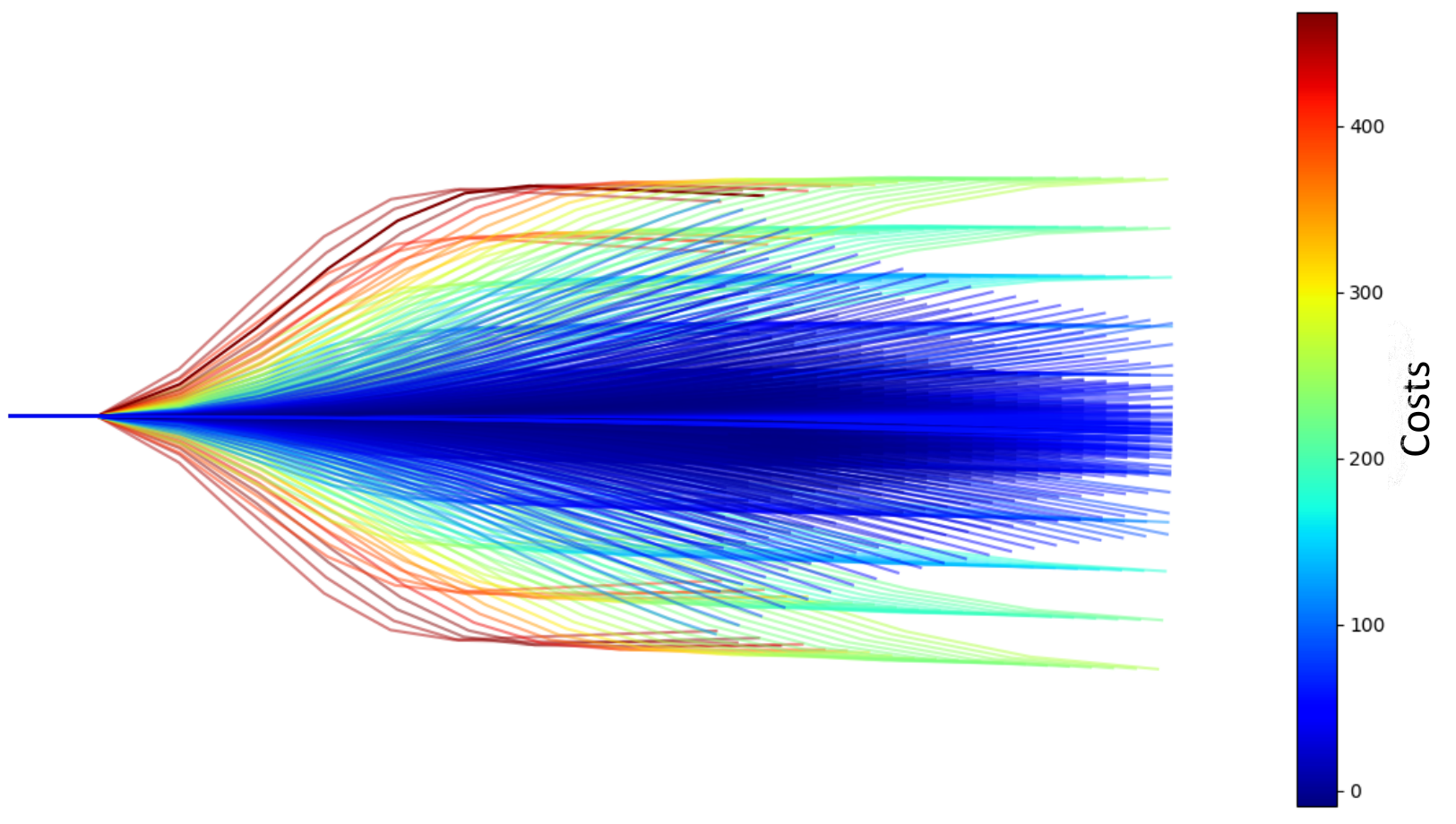}\\
    (a) {} & 
    (b) \\
    \end{tabular}
 \caption{ \textbf{(a):} The predicted trajectories for all the actors in the scene obtained via the ILVM \citep{casas2020implicit} network. The color gradient shows different timesteps in the predicted trajectory. White curves illustrate the ground truth behavior of the actor. \textbf{(b):} Control trajectories in continuous (x,y) frame generated from the baseline Polynomial Frenet method over a straight path. The color gradient from blue to red indicates the change in costs for the control trajectories considering dynamics and lane following penalties. }
    \label{fig:ilvm_planner_cost}
\end{figure*}

An AV's observation and action history $h_t$ is high dimensional, consisting of a historical sequence of sensor observations, map states, and vehicle states. Starting from this raw data, we seek to construct a context feature vector representing the belief state $b(h_t)$ for conditioning the flow network. In this work, we use a pretrained detector \cite{yang2018pixor} which takes as input a voxelized LiDAR point cloud and rasterized map state and constructs an internal feature representation of $b(h_t)$, which we denote as  $\overline{b}(h_t)$. Output \somi{detections} \hs{what does output states mean?} from this detector head are forwarded to a prediction head (ILVM \cite{casas2020implicit}) for generating scene predictions. ILVM is a graph neural network used for generating multimodal future actor distributions. These actor trajectory predictions are passed to the \somi{ Should we switch this white-box planner ? I think we should stick to 1 name here.}trajectory planner cost functions while the internal feature vector representation $\overline{b}(h_t)$ is used by the flow module as a conditioner. An illustration of the ILVM output can be seen in Figure \ref{fig:ilvm_planner_cost}(a).

% \begin{equation*}\label{eq:backbone}
%     \overline{b}(s) = backbone(h_t)
% \end{equation*}

% The future actor predictions from this network, which form the interpretable representation $b(s)$, are used for costing samples by a white box planner as described in section \ref{section:traj_planner}.

% \begin{wrapfigure}
%   \includegraphics[width=0.5\linewidth]{figures/beliefs/belief1.png}
% \caption{Interpretable belief representation forms the context vector which is input to the flow network and used for sampling trajectories.}
% \end{wrapfigure}

% \begin{figure}
%     \centering
%  \includegraphics[width=0.5\linewidth]{figures/beliefs/belief1.png}
%  \includegraphics[width=0.5\linewidth]{figures/beliefs/belief2.png}
% \caption{Interpretable belief representation forms the context vector which is input to the flow network and used for sampling trajectories.}
% \label{fig:flow_vs_plt}
% \end{figure}
 
\subsection{Trajectory Planner}\label{section:traj_planner}
%  such that the optima represents safe, comfortable, and socially responsible behavior.

The purpose of a trajectory planner in an AV system is to find control policy $\mathbf{\pi}^*$ corresponding to the optima of the cost manifold from  Eq. \ref{eq:optimal_pi_finite_horizon}. In this work, the planner comprises of two parts: a control-trajectory sampling scheme and an interpretable whitebox costing module. \somi{The output of the planner is the deterministic control policy providing minimum expected cost}. The sampling scheme is our main contribution and discussed in section~\ref{section:sampling_policy}. 

The costing module is a linear combination of a number of cost functions, encoding preferences for safety, performance and user comfort. \somi{We provide detailed descriptions of the cost functions in Appendix \ref{ap:costs}.} We utilize the costing module in two ways: During offline training as a supervision to learn a stochastic sampling policy, and in online testing to select the best trajectories for execution. The module ingests map data, vehicle platform state, and probabilistic multimodal trajectory predictions for other actors future states to generate a scalar expected cost. We rely on dubins model to simulate forward dynamics of the AV. A key requirement is that the cost functions and dynamics used for forward propagation to be differentiable to support training.

% \begin{figure}
%   \centering
%   \includegraphics[width=1\linewidth]{figures/cVAE Daigram.jpeg}
%   \caption{Conditional Variational Autoencoder}
%   \label{fig:cvae}
% \end{figure}

\subsection{Expert Demonstration Encoding}\label{section:vae}

We construct an encoding space of human expert control trajectory demonstrations using a conditional Variational Autoencoder (cVAE). The cVAE learns a lower dimensional subspace of human-like trajectories using a large dataset of human demonstrations.
% Figure \ref{fig:cvae} shows simplified version of our cVAE.
The inputs to the VAE are $x$ which is the human expert control trajectory and a condition vector $c$ which consists of a fixed-length history of AV control trajectory. The cVAE is trained following the $\sigma$VAE ~\cite{rybkin2020sigmavae} method which allows for the weight between the MSE and KL divergence terms in the loss function to be learned removing the need of additional hyperparameter tuning.
\begin{equation}\label{eq:sigma_vae}
    \mathcal{L} = D \ln{\sigma} + \frac{D}{2 \sigma} MSE(\hat{x}, x) + D_{KL} (q(z|x) || p(z)).
\end{equation}

 The $\sigma$VAE is trained separately on all human expert control trajectories from the dataset and is kept frozen during the flow training. During the main training loop only the decoder is used to decode trajectory samples from the latent space.

% \textcolor{green}{cVAE is trained with convolutional layers for the encoder and decoder, batch normalization, and ReLu activation functions. <-- Maybe move to experiments}

% \begin{figure}
%   \centering
%   \includegraphics[width=1\linewidth]{figures/autoregressive_flow.jpeg}
%   \caption{General Autoregressive Flow Architecture}
%   \label{fig:autoregressive_flow}
% \end{figure}
\subsection{Sampling Policy}
\label{section:sampling_policy}

We propose to learn a stochastic sampling policy in the latent space of the cVAE described above. \st{Towards this goal,} We formulate the cost functions as an energy based model (Eq.\ref{eq:energy_function}) and \somi{learn} \st{learning} a maximum entropy policy minimizing the reverse KL divergence to it. We utilize Neural Autoregressive Flow (NAF) to facilitate efficient learning of multimodal energy landscape induced by the cost as well as to obtain concrete probability estimates which provides a score for each sampled trajectory. The affine transformations which were used in earlier flow models such IAF \cite{kingma2016improved} and MAF \cite{Papamakarios2017MaskedAF} supported efficient inversion and log determinant calculation required for (\ref{eq:flow_log_density}) but are not as flexible in representing multimodal distribution as NAF as shown in \cite{huang2018neuralaf}. Our NAF policy will take as input a vector sampled from a prior distribution ($z_0 \sim  \mathcal{N}(0, 1)$) and the belief state of the AV ($\overline{b}(h_t)$) as a conditioner. It outputs $z_N$ in the latent space of our VAE, as a result of a number of flow transformations along with its probability.

In this work, we aim to learn a mapping from a known prior distribution, $q(z_0) = \mathcal{N}(0,1)$, to the target distribution defined by the planner cost surface in Eq. \ref{eq:energy_function}. We formulate the mapping as the optimization as a reverse-KL divergence minimization:

\begin{equation}\label{eq:kl_divergence_loss}
    \text{argmin}_{\theta}D_{\text{KL}}\left[q(z_N |\theta, b)~||~ J(z_N | b) \right] 
\end{equation}

where $q(z_n |\theta, b)$ is the output of the flow model in Eq. \ref{eq:flow_log_density} and $J(z_n | b)$ is the likelihood of that output under the planner cost surface.
We train the normalizing flow policy by obtaining the scene context feature vector from the backbone network as described in section \ref{section:scene_conditioning} and drawing L samples from the prior distribution $z_0 \sim \mathcal{N}(0, 1)$. \somi{FIX THIS The per sample loss of the flow mapping minimizes Eq. \ref{eq:kl_divergence_loss} with the output of Eq. \ref{eq:naf} according to}

\begin{equation}\label{eq:prob_density_matching_loss}
    loss = -\log ( J(z_N | b)) -  \sum^{N}_{n=1} \log \det |\frac{dz_n}{dz_{n-1}}|
\end{equation}
Here, we can ignore the partition function as it does not depend on the parameter $\theta$. \somi{We elaborate more on the partition function in Appendix \ref{ap:partition_function}.} The obtained solution can be interpreted as a maximum entropy sampling distribution for the whitebox planner. A maximum entropy policy can be proved to be a solution of robust-reward control problem in the presence of an adversary as shown in \cite{eysenbach2019if}. Even in the setting without an adversary, the adversarial objective bounds the worst case performance of the agent. This is similar to the policies obtained from state of the art model-free Reinforcement Learning method  SAC~\cite{haarnoja2018soft}. The solution to the planning problem (Eq.\ref{eq:optimal_pi_finite_horizon}) is given by the maximum aposteriori estimate \somi{(MAP)} under the learned policy parameterized by $\theta$.

\begin{center}
\footnotesize
\label{tab:motion_planning_metrics}
\begin{tabularx}{1.0\columnwidth} { 
  | >{\raggedright\arraybackslash}X 
  | >{\centering\arraybackslash}X
  | >{\centering\arraybackslash}X
  | >{\centering\arraybackslash}X 
  | >{\centering\arraybackslash}X 
  | >{\centering\arraybackslash}X 
  | >{\centering\arraybackslash}X 
  | >{\centering\arraybackslash}X 
  | >{\raggedleft\arraybackslash}X | }
 \hline
 Models & Avg. Jerk ($m/s^3$) & Avg Lat Accel ($\text{rad}/s^3$) & Avg. Progress (m) & Collision 1.5s & Collision 2.0s \\
 \hline
 \hline
  Human  & 4.14  & 1.93 & 13.05  & 0.00 & 0.00 \\
 \hline
 \hline
 Frenet  & 3.03  & 3.89 & 13.27  & 0.01 & 0.02 \\
 \hline
 NAF  & 1.59  & 4.10 & 10.44 & 0.03 & 0.11 \\
 \hline
%  NAF + $\sigma$-VAE + IL   & 1.53  & 3.05 & 13.21 \\
%  \hline
 $\sigma$-VAE  & 3.47  & 3.20 & 13.29  & 0.00 & 0.1 \\
 \hline
 FlowPlan  & 1.53  & 3.05 & 13.21 & 0.00 & 0.1 \\
 \hline
\end{tabularx}
\caption{Motion Planning Metrics: \hs{TODO}}
\end{center}

\begin{figure}
\vspace{1mm}
    \centering
    \begin{tabular}{cc}
    \includegraphics[width=0.48\textwidth]{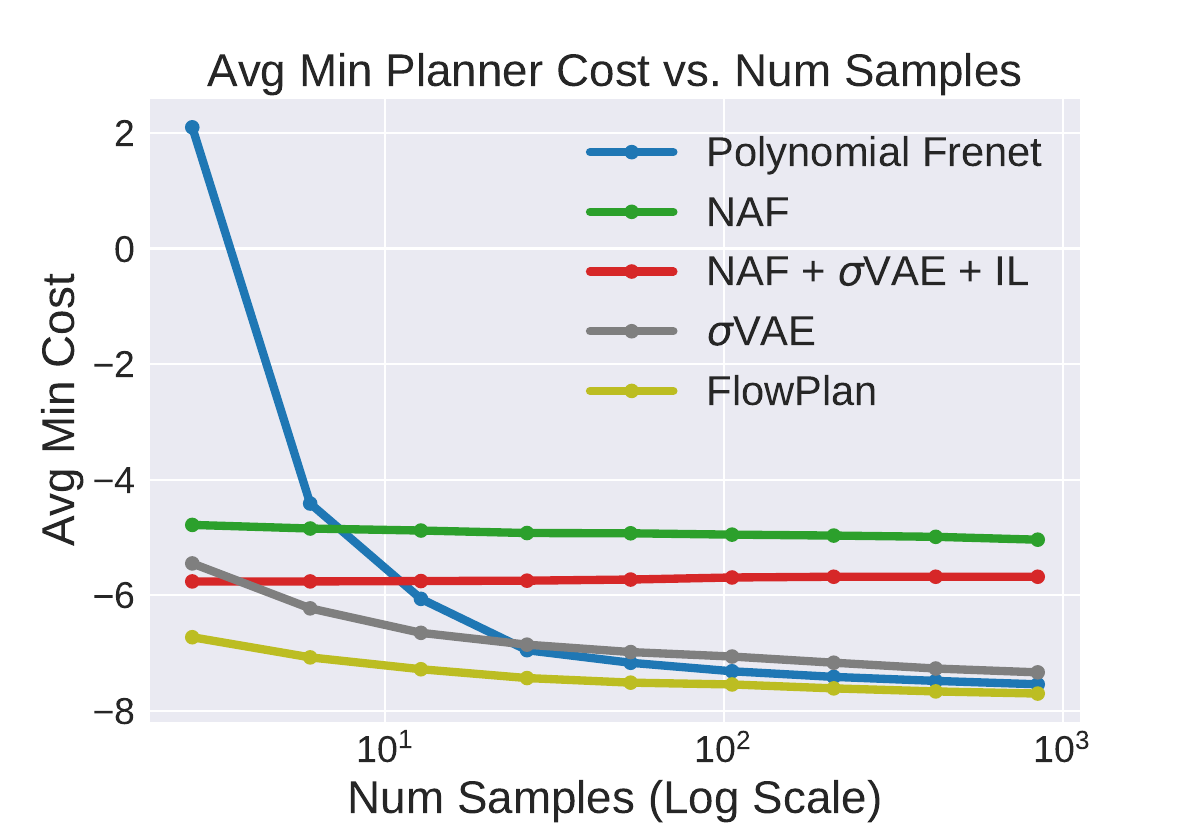}
    & \includegraphics[width=0.48\textwidth]{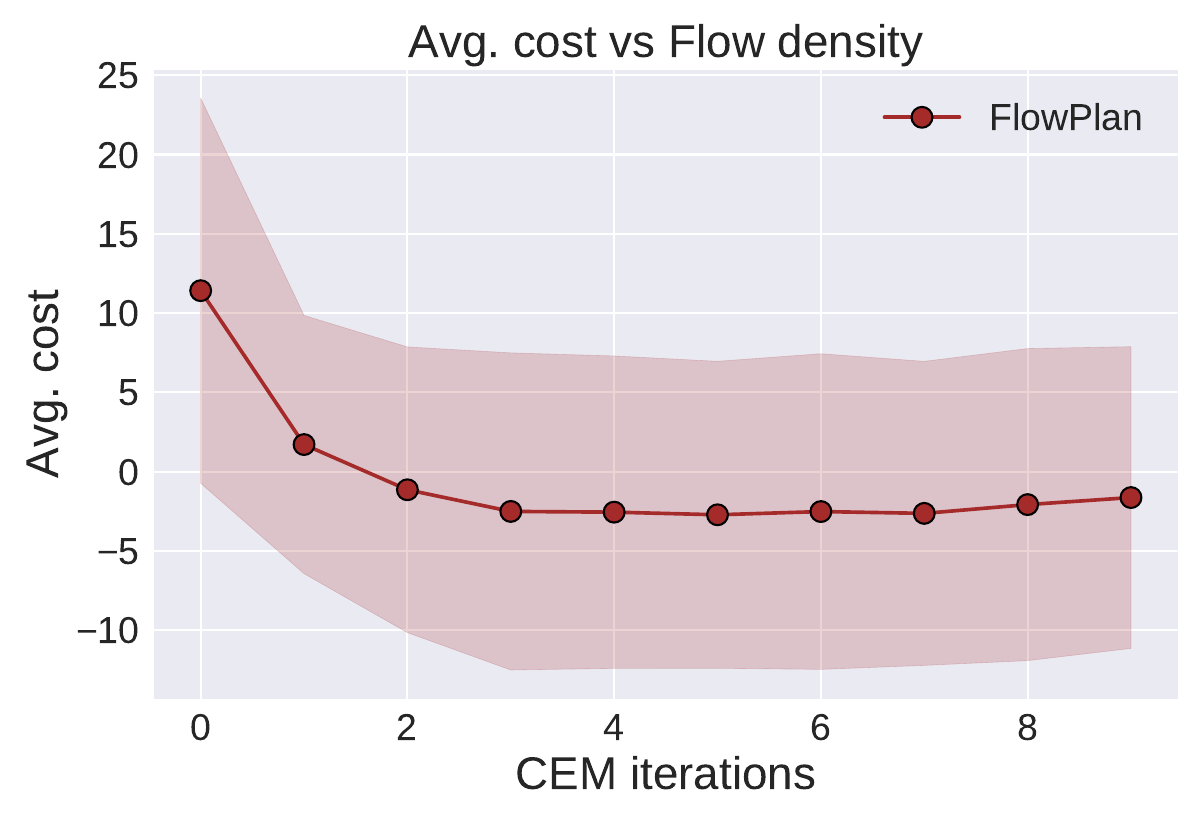}\\
    (a) {} & 
    (b) \\
    \end{tabular}
 \caption{ \textbf{(a):}  A comparison of different sampling techniques used for generating low cost control trajectories. We measure the average cost of the best performing control trajectory for every scene in the evaluation set. Our method FlowPlan outperforms the baseline Polynominal Frenet method especially in the low sample regime. \textbf{(b):} We demonstrate that high probability regions of the NAF output distribution correspond to low cost surface regions using cross-entropy method (CEM). As CEM iterations increase the corresponding average cost under the planner cost surface of all samples in the CEM set decreases. }
    \label{fig:flow_vs_plt}
\end{figure}

\subsection{Target Distribution Learning}
\label{subsec:target_distribution_learning}
% \begin{wrapfigure}{r}{0.4\textwidth}
%   \centering
%   \includegraphics[width=1.0\linewidth]{figures/pdf_figures/cem_iter.pdf}
%   \caption{Planner costs goes down with increasing CEM iterations and increase flow density.}
%   \label{fig:CEM}
% \end{wrapfigure}

In this experiment, we are interested in empirically verifying if high probability density regions of the FlowPlan output distribution correspond to low planner cost. We propose finding high-probability regions in the output distribution of FlowPlan using the cross-entropy method (CEM). In CEM, we sample $n$ times from an initial sampling distribution. The top $e$ samples with highest probability density under output distribution from FlowPlan are selected and used to update the mean and the variance of the original sampling distribution. After $N$ iterations of refinement we output the mean of the resulting sampling distribution as our latent variable which has the maximum density under the output distribution.
\begin{equation}
\begin{split}
        A_i = & \ \{z^i\}, A_i \sim \mathcal{N}(\mu^m,\,\Sigma^m) \, \forall i \in n \\
        & A_{\text{elites}} =  \  \text{sort}(A_i)[-e:] \\
        & \mu^{m+1} =  \ \alpha * \text{mean}(A_{\text{elites}}) + (1-\alpha)\mu^m\\
        &\Sigma^{m+1} =  \ \alpha * \text{var}(A_{\text{elites}}) + (1-\alpha)\Sigma^m
\end{split}
\label{eq:CEM}
\end{equation}

In Figure \ref{fig:flow_vs_plt} (b), we show that with each iteration of CEM we sample higher probability trajectories in the FlowPlan output distribution and on evaluation of these trajectories we find the average planner cost decreases. This shows that high probability control trajectories under our learnt distribution correspond to low costs in the planner cost manifold. We can further use this method to improve our performance as shown in Appendix \ref{ap:flow_resampling}.
%===============================================================================

\section{Discussion}
\label{sec:discussion}

We present FlowPlan, a normalizing flow approach for generating control trajectory samples along with associated probability density under the planner cost surface for AVs. As the flow model is connected to a learned perception \& prediction model which generates interpretable motion forecasting, the model leverages the full scene context during inference and adds little computational overhead to the existing AV stack. We compare this model using a dataset consisting of real world driving examples and show this approach is per sample more efficient to alternative approaches. 

\textbf{Limitations and Future Work.} Because the model learns a non-transparent mapping from the prior distribution to the target, in order to ensure that safety maneuvers, such as max braking, are always in the considered trajectory set these have to be added in through an outside process. Additionally, the SDV trajectory samples are generated independently from the motion forecasting of other actors the predicted actions of other actors in the scene are not conditioned on the SDV intent. In the future we would like to extend this work to a unified probabilistic generative model that samples the SDV trajectory samples jointly with the motion forecasts of other actors.

%===============================================================================

% The maximum paper length is 8 pages excluding references and acknowledgements, and 10 pages including references and acknowledgements

\clearpage
% The acknowledgments are automatically included only in the final version of the paper.
% \acknowledgments{If a paper is accepted, the final camera-ready version will (and probably should) include acknowledgments. All acknowledgments go at the end of the paper, including thanks to reviewers who gave useful comments, to colleagues who contributed to the ideas, and to funding agencies and corporate sponsors that provided financial support.}

%===============================================================================

\bibliography{example}  % .bib

\onecolumn
\appendix

\section*{\Large Supplementary Material for Imitative Planning using Conditional Normalizing Flow}

\subsection{Experiment Details}
\label{ap:experiment_details}

\emph{Expert Dataset:} HES-4D contains more than one million frames collected over several cities in North America with a 64-beam, roof-mounted LiDAR. The labels are precise 3D bounding box tracks with a maximum distance from the self-driving vehicle of 100 meters. There are 6500 snippets in total, each 25 seconds long. We have access to high definition maps capturing the geometry and the topology of each road network in every city. Following previous works in joint perception and motion forecasting \cite{casas2018intentnet,casas2019spatially} we consider a rectangular region centered around the self-driving vehicle that spans 144 meters along the direction of its heading and 80 meters across. This dataset involves trajectories observed in various situations like Lane Keeping, Merging, Intersections among others. Each trajectory is trimmed to 4 second blocks.

\subsection{Model Details} \label{section:model_details}
\emph{$\sigma$-VAE:} We use a conditional variational auto-encoder to compress the trajectory to a small latent space of 5 dimensions. This is motivated by the fact that the trajectories feasible under the kinodynamic constraints of the AV are limited and lie in a much smaller latent space. The context used for the conditioning is the 2 second history and is used to reconstruct the trajectory for the other 2 seconds of the trajectory obtained from the step above. Note that trajectories in these case are represented as control inputs of acceleration, steering pair and not the position-angle form. Rather than hand-tuning a desired weight between reconstruction error and KL divergence with prior in the VAE loss, we use $\sigma$-VAE which allows for this tuning to happen automatically. For the encoder and decoder, we use 3 convolutional layers with batch normalization followed by 2 fully connected layers.

\emph{Normalizing flow:} We use the deep sigmoidal flow variant of the Neural Autoregressive flow \cite{huang2018neuralaf} in this work. Our flow module comprises on 3 fully connected layers with 256 neuron units with exponential linear unit(elu) non linearities. The latent space of 5 dimensions obtained after passing the input through the encoder is transformed into a multimodal latent sampling distribution for low cost trajectories.

\subsection{Imitation Learning Architecture}
In this section we present the imitation leaning (IL) architecture used for experiments in section \ref{subsec:planning_performance}. The exact details of the models is exactly same as described in section \ref{section:model_details}. Our IL framework uses different model architectures during training and evaluation.

\begin{wrapfigure}{r}{0.5\textwidth}
    \begin{center}
    \includegraphics[width=1\linewidth]{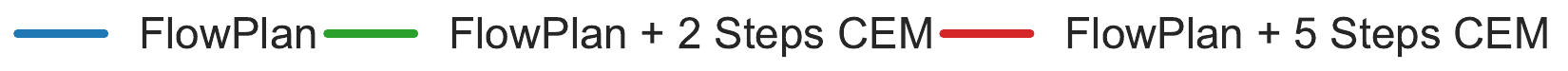}
    \includegraphics[width=1\linewidth]{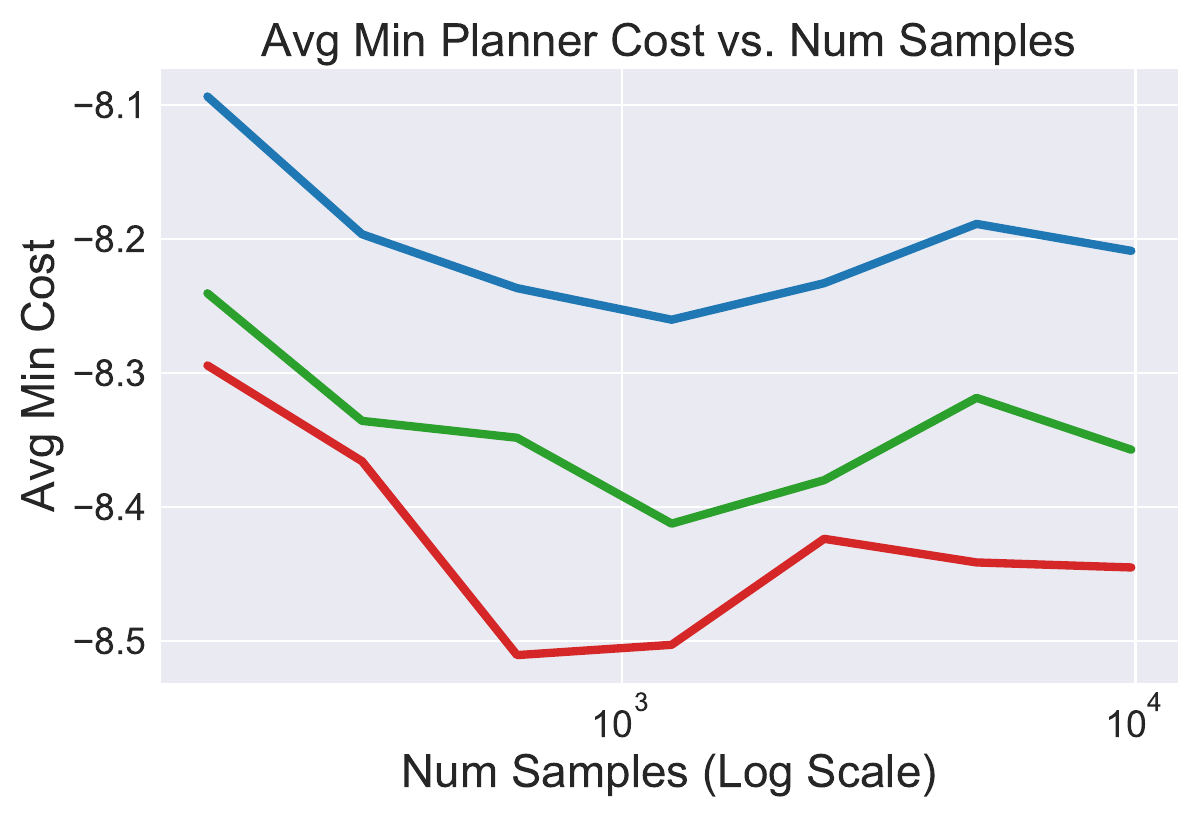}
    \end{center}
    \caption{Planner cost decreases with increase in CEM iterations on FlowPlan.}
    \label{fig:cem_flow}
\end{wrapfigure}

\subsubsection{Training} The goal here is to learn from expert (human) demonstrations given the scene context. Since we are not costing any trajectories we do not need any detections, prediction and differentiable costing modules. We pass the expert trajectory through the cVAE Encoder, conditioned on AV history, to model the expert trajectory in lower dimension latent embedding referred to as $Z_{N}$. In reference to NAF, Encoder encodes the expert trajectory in a complex (multimodal) distribution. Forward NAF, conditioned on scene condition, maps the $Z_{0} = NAF(Z_{N})$ to a normal distribution. We train this model using maximum log-likelihood loss.

\subsubsection{Evaluation} The goal here is sample trajectories of what most likely human would do given the scene condition, as compared to FlowPlan where goal is sample trajectories which minimize the cost functions. The architecture here is same as FlowPlan except the NAF model used here is inverse of NAF model used for training, $Z_{N} = NAF^{-1}(Z_{0})$. Here the flow model provides a mapping from normal distribution to a complex distribution, conditioned on the scene. We pass the $Z_{N}$ through cVAE Decoder to obtain trajectories from the latent variable. We cost the trajectories with a differentiable costing module to find the best trajectory under planner cost surface for the given scene.

\begin{figure}[t]
  \centering
  \includegraphics[width=1\linewidth]{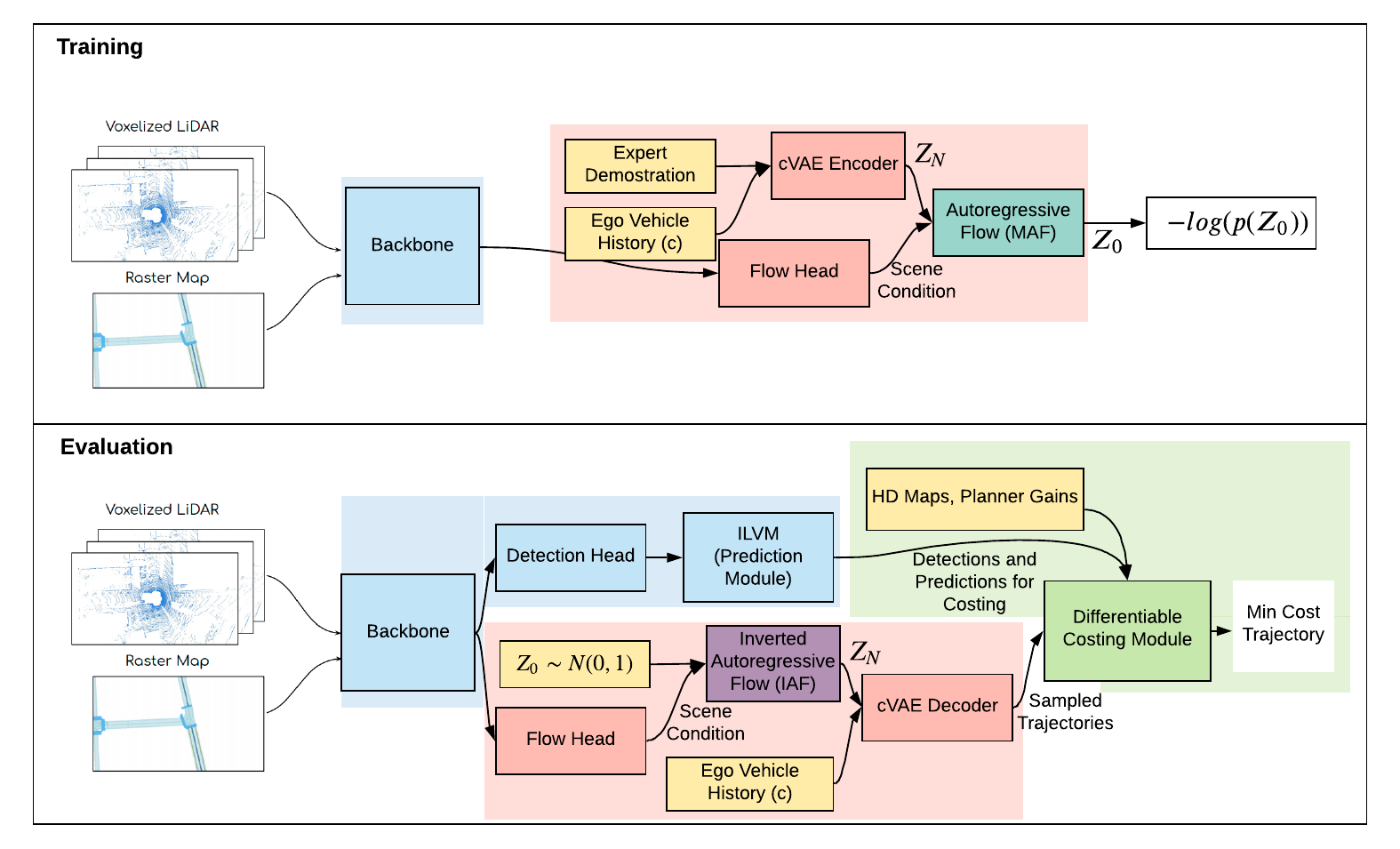}
  \caption{Imitation Learning Architecture.} 
  \label{fig:il_architecture}
\end{figure}

% \clearpage

\section{Improving Flow performance using resampling}
\label{ap:flow_resampling}
FlowPlan provides us with a good initial sampler producing trajectories which have low costs. Generally planners on AV work under time constraints, and have a fixed amount of available time budget for planning. For example, a typical planner aims to produce plans at 10 Hz frequency.

In this section we study improving upon the already obtained plans when we have additional time budget available. 

To achieve this we rely on a similar method as used in Section \ref{subsec:target_distribution_learning}. We use CEM to refine our solutions iteratively. We sample initially using FlowPlan and successively improve upon the solution using the Planner Cost (in contrast to the Flow density) as the CEM objective now. We experiment with 2 and 5 iterations of CEM and find that increasing CEM iterations increases the quality of solution. For a fair comparison against the basic FlowPlan we sum up the number of samples used during all the CEM iterations (x-axis of Figure \ref{fig:cem_flow}). 

% \subsection{CEM on big eval}
% \subsection{Dynamics Bounds}

\section{$\sigma$-VAE Latent Space}
\label{ap:sigma_vae_latent_space}
In this section we share some results which give more insights into our $\sigma$-VAE's latent space.
\subsection{Gaussian Sampling}

\begin{figure}[h]
  \centering
  \includegraphics[width=1\linewidth,height=5.5cm]{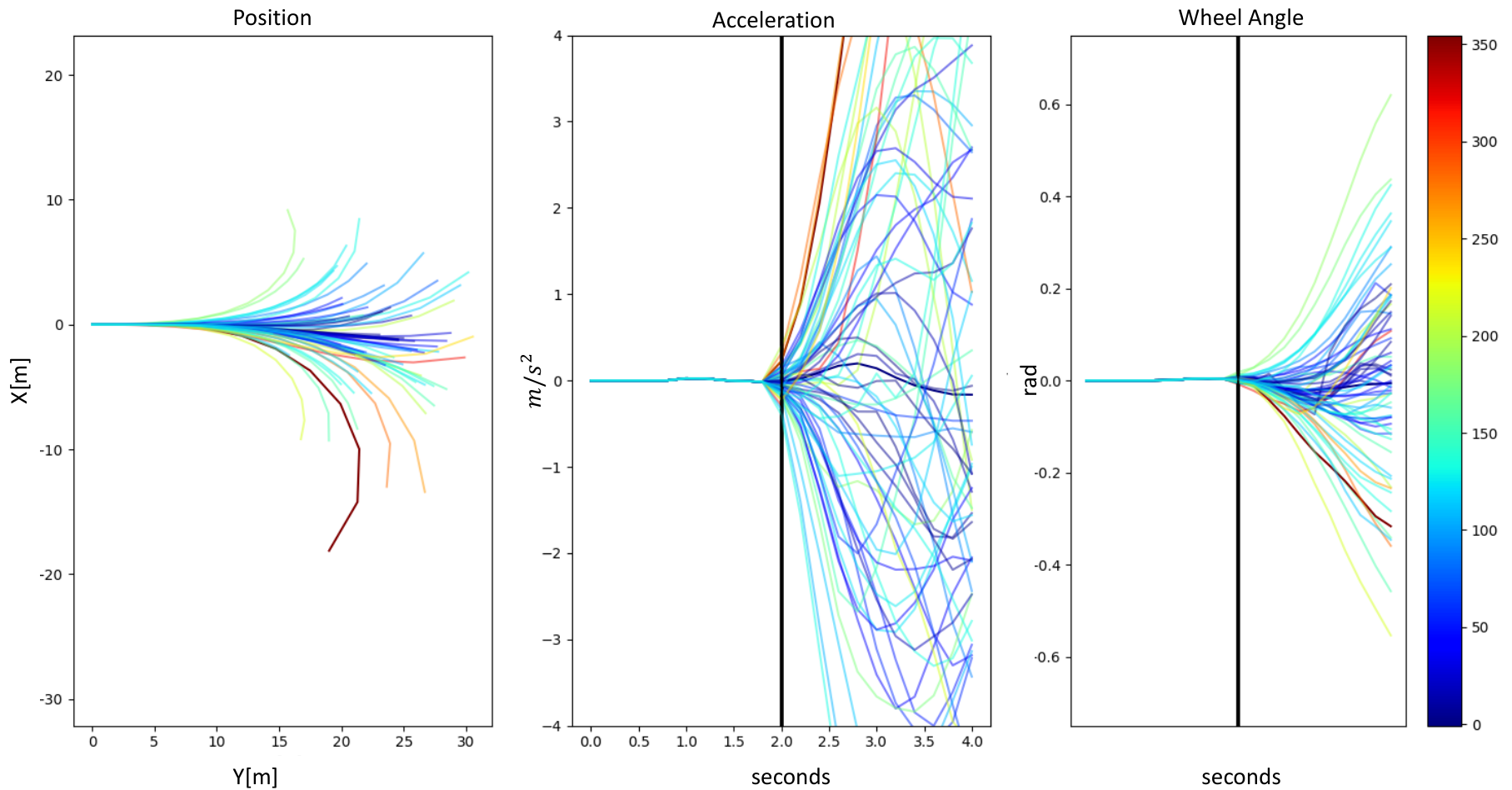}
  \includegraphics[width=1\linewidth,height=5.5cm]{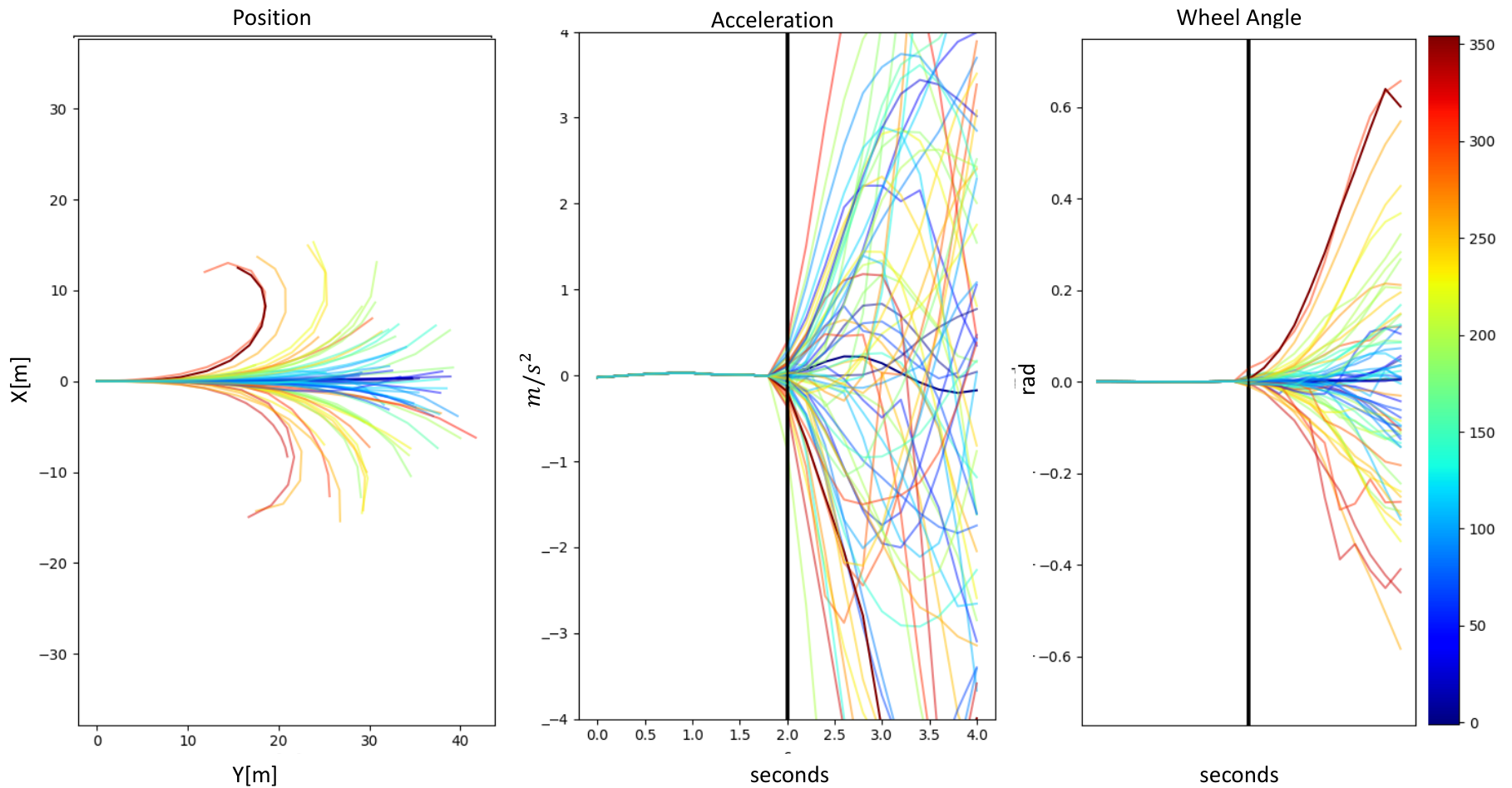}
    \includegraphics[width=1\linewidth,height=5.5cm]{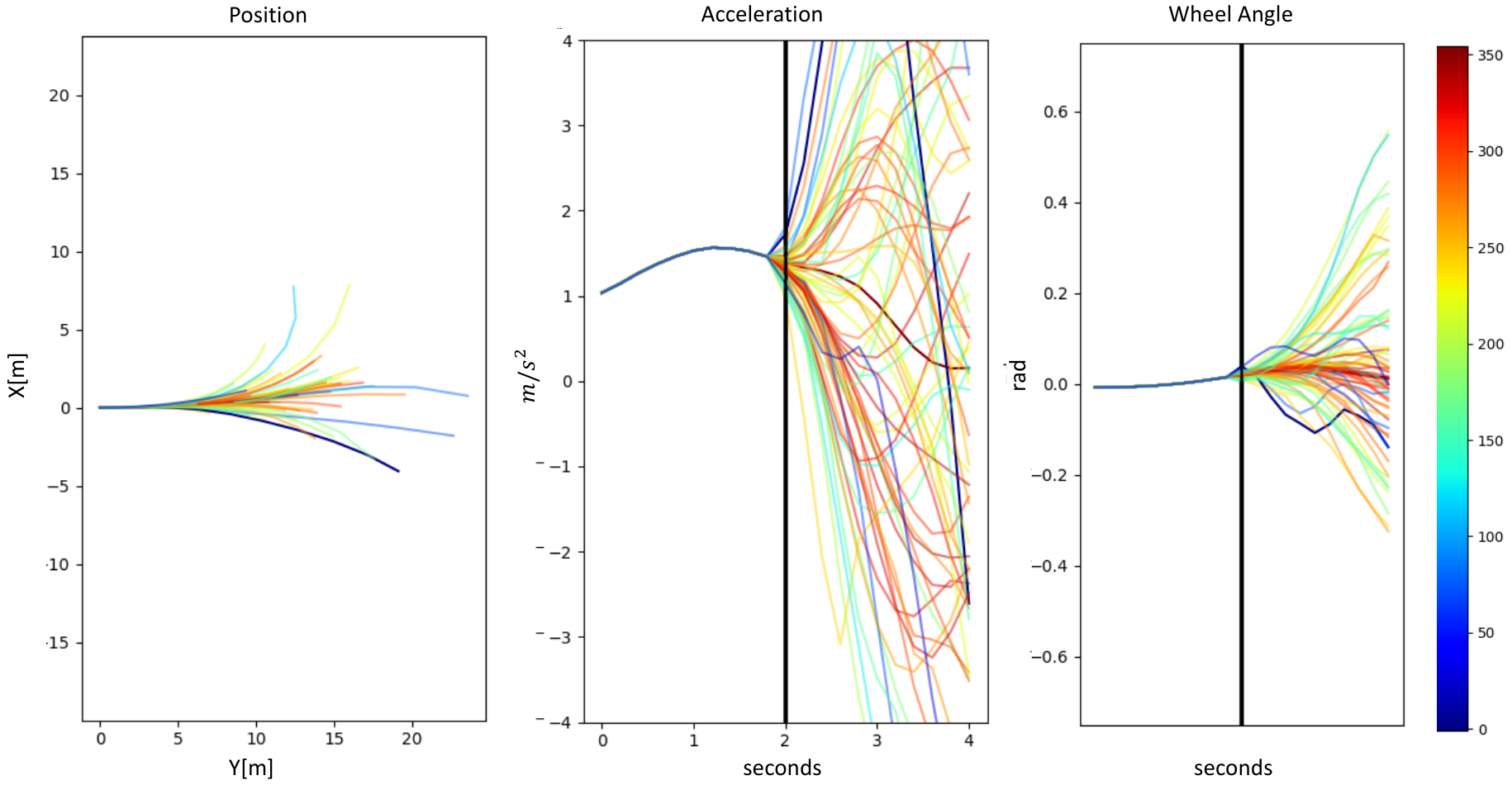}
  \caption{Output of $\sigma$-VAE when sampling from $N(0, 1)$ from the latent space. This demonstrates that naively sampling from $N(0, 1)$ results in control trajectories with high coverage of the action space but not necessarily in low cost trajectories under the planner cost surface. Colors here illustrate the respective cost of the control trajectory.} 
  \label{fig:gauss_samples}
\end{figure}

We learn a latent embedding using a diverse set of trajectories obtained from expert demonstrations. Figure \ref{fig:gauss_samples} shows some examples fo the trajectories learned by the VAE. It also shows that the trajectories produced by the VAE may not be low cost and can be compared to Figure \ref{fig:good_flow} where we see that flow learns a tight low cost distribution in this latent space. 

\subsection{Latent Space Interpolation}

\begin{figure}[h]
  \centering
    \includegraphics[width=1\linewidth,scale=1]{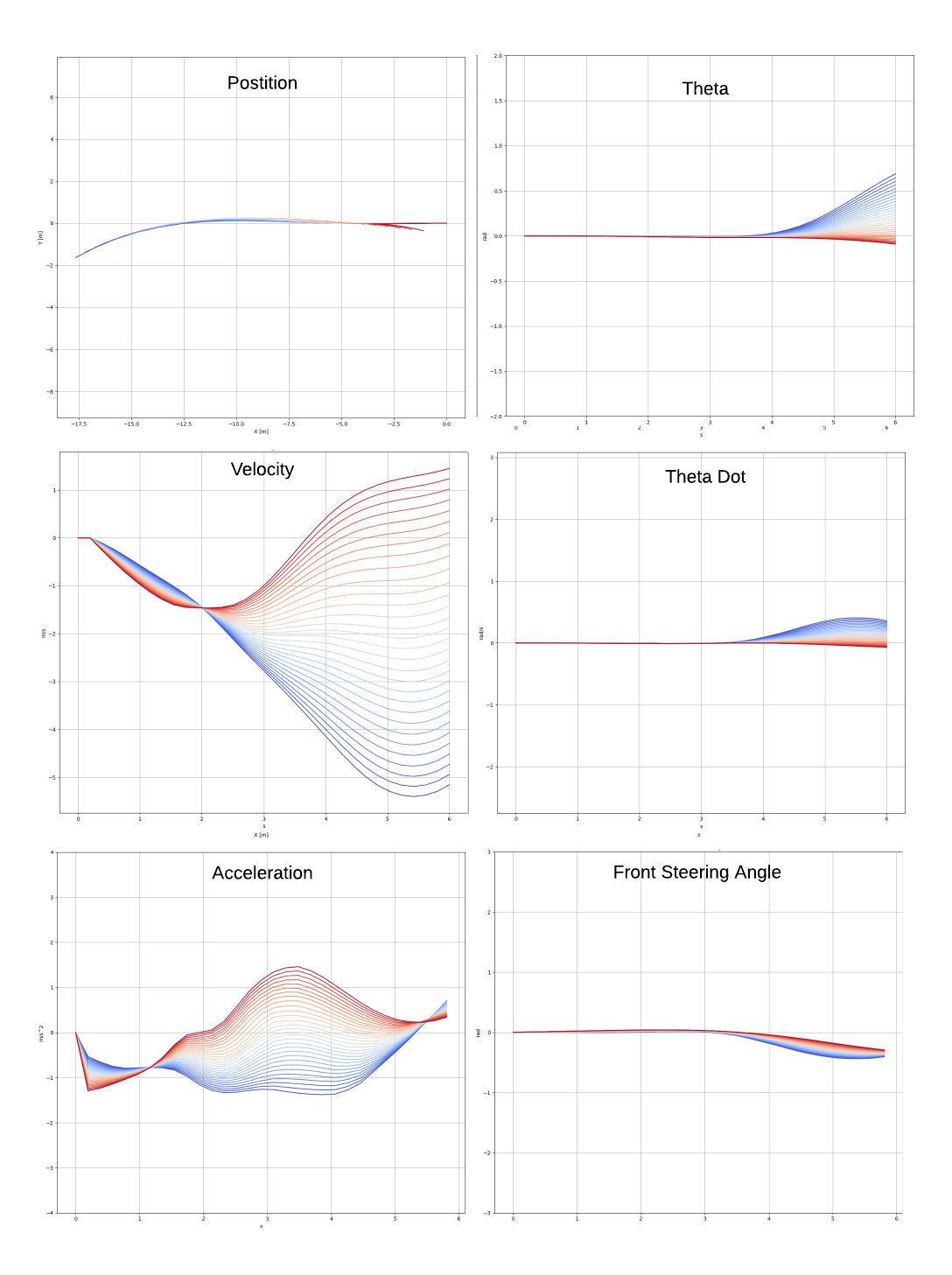}
  \caption{Smooth trajectories from $\sigma$-VAE's latent space. Here we linearly interpolate between latent variables, from red to blue, belonging to the actual dataset and show that the new latent variables result in smooth output. Theta is the heading of the AV.} 
  \label{fig:interpolate}
\end{figure}

Figure \ref{fig:interpolate} shows the trajectories we obtain when interpolating in the latent space. 

\clearpage
\subsection{$\sigma$-VAE Latent Space Analysis}
\begin{wrapfigure}{r}{0.5\textwidth}
    \begin{center}
    \includegraphics[width=1\linewidth]{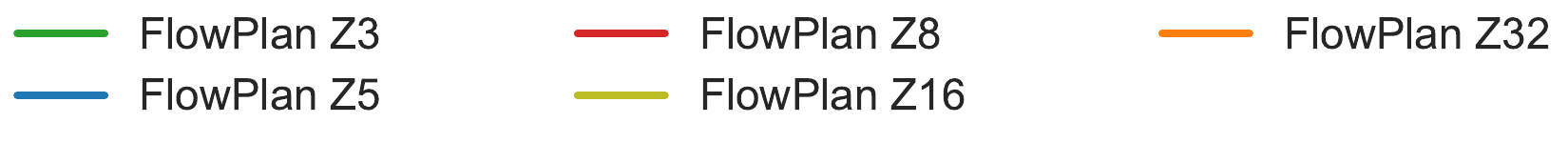}
    \includegraphics[width=1\linewidth]{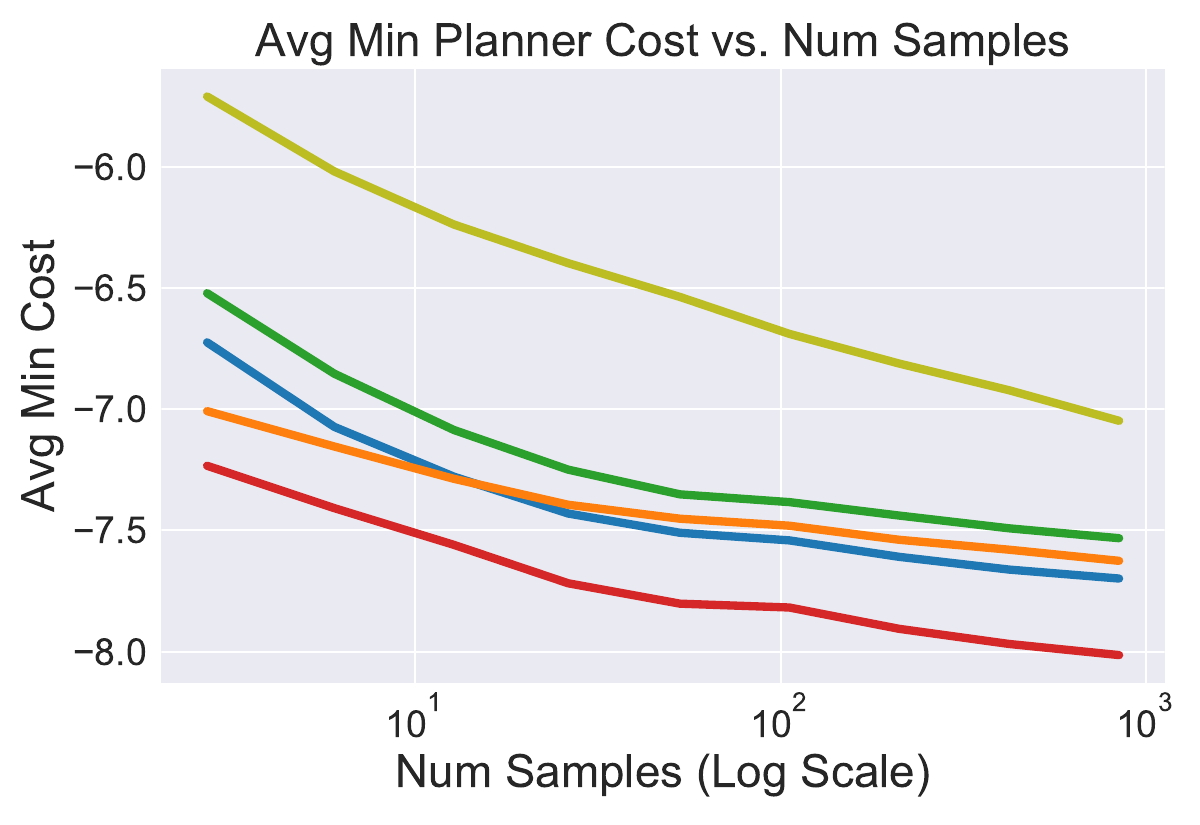}
    \end{center}
    \caption{Latent Dim Comparison} 
    \label{fig:latent_analysis}
\end{wrapfigure}

In this section we experiment with different latent dimensions for FlowPlan. A smaller dimension makes it harder for the latent space to capture the high dimensional trajectory distribution and a high dimensional latent space increases the complexity of learning and generalization. We find that a latent dimension of 8 achieves the minimum average cost when tested on the HES-4D dataset. In Figure \ref{fig:latent_analysis}, we see that the average cost of FlowPlan decreases as we increase the latent dimension from 3 to 8 and then increases as the dimensions are increased to 32.

\section{Partition Function}
\label{ap:partition_function}
KL divergence is a good metric to evaluate if the sampling policy we learn is close to the Boltzmann distribution induced by the cost surface. We compute the KL divergence using a sample based estimate, but we need a partition function since our cost functions are unnormalized. We rely on importance sampling to compute a sample based estimate of the partition function, where the importance sampling distribution is our current flow distribution. The importance sampling distribution becomes optimal as the policy approaches the Boltzmann distribution induced by the cost functions.

\begin{equation}
    \begin{split}
        Z &= \int e^{-C(\tau)} d\tau \\
        \hat{Z}&= \E{z \sim Z_N}{\frac{e^{-C(decoder(Z_N))}}{p(Z_N)}} 
    \end{split}
\end{equation}

where C is the unnormalized cost function, which costs the entire trajectory following Appendix \ref{ap:costs}.

A sample based estimate for the reverse-KL divergence can be obtained as follows:

\begin{equation}
    \begin{split}
        D_{\text{KL}}[q(z_N|\theta, b)~||~J(z_N|b)] &= \E{z_N \sim Z_N}{\log q(z_N|\theta, b) - \log \frac{e^{-C(decoder(Z_N))}}{Z}}\\
        &= \E{z_N \sim Z_N}{\log q(z_N|\theta, b) + C(decoder(Z_N)) + \log \hat{Z}}
    \end{split}
\end{equation}
The decoder takes the latent embedding of a trajectory and outputs the reconstructed trajectory.

\section{Cost Functions}
\label{ap:costs}
In this work we generate control trajectories in the space of steering angle and acceleration. These control trajectories will be simulated by a forward bicycle model to obtain trajectories in \textit{frenet} space ~\cite{werling2010optimal} and Cartesian space. Frenet space represents the trajectory of a car as latitudes and longitudes based on a nominal path that a car is expected to follow at each points. We will consider trajectories in both frames and the predictions obtained for each actor in the scene as a interpretable belief for costing purposes. 

We use a set of costs that allow for safe driving with user comfort in mind. In particular our cost functions can be divided into five costs: path distance, centerline, obstacle collision, jerk, and twist costs. We expand upon all the cost functions in detail below:

\subsection{Path Distance Cost $C_d$}
A basic objective of the car is to move along the directed centerline. A \textit{centerline} is defined as a nominal path safe to follow in presence of no obstacles. It is obtained as a part of the High Definition maps and in this work is just the centerline equidistant from the two lane boundaries. We reward the agent to cover as much distance on the centerline as possible. The cost function for the distance cost looks as follows:

\begin{equation}
C_d = - \text{Agent cartesian displacement projected on the centerline}  
\end{equation}

\subsection{Centerline Cost $C_c$}
In addition to travelling as much longitudinal distance as possible we would like our AV to stay close to the centerline. We do this be penalizing the normal displacement of a cartesian trajectory to the centerline along each control point along the trajectory. 
\begin{equation}
    C_c = \sum_{t=1}^T d_t^2
\end{equation}
where $d_t$ is the normal distance (latitude) of the $t$ timestep in trajectory to the centerline.

\subsection{Obstacle Collision Cost $C_o$}
Penalizing collisions is an important aspect to ensure safety of AV. We consider the future actor predictions in the scene and unrolled trajectory of the AV using a dynamics model to check for possible collisions. The cost function is summed over all the probabilistic elements of the scene using the probabilities output by the prediction module. The expected cost is passed through a rectified linear unit to penalize cost for only those actors that enter a 3 meter radius of the car. 
\begin{equation}
    C_o = \sum_{obstacles} \text{prob}_{obs} * (rectified(\text{distance to obstacle}))^2
\end{equation}

\subsection{Jerk Cost $C_j$}
Jerk is defined as rate of change of acceleration. Minimizing jerk is crucial to obtain a comfortable user experience. We encode this directly in our cost functions. 

\begin{equation}
    C_j = \sum_{t=1}^{T-1} (\dot{a})^2
\end{equation}

where $a$ is acceleration and $\dot{a}$ is the jerk (first derivative of acceleration).

\subsection{Twist Cost $C_t$}
Ensuring smooth changes in curvature for the trajectory is another aspect of encoding user preference for comfort. We do this by directly imposing a smoothness constraint on the steering of the vehicle. 

\begin{equation}
    C_t = \sum_{t=1}^{T-1} \dot{c}_i^2
\end{equation}
where $c$ is curvature and $\dot{c}$ is the twist (first derivative of curvature).

\subsection{Cost Gains}
The final cost for a trajectory is the weighted combination of the cost functions applied to the trajectory by cost gains. The gains are manually tuned for best performance and interpretability.

\begin{equation}
    \textbf{Final Cost} = w_d*C_d + w_c*C_c + w_o*C_0 + w_j*C_j + w_t*C_t
\end{equation}

% \section{Multi-modal Cost Surface}

% \begin{figure}[h]
%   \centering
%   \includegraphics[width=1\linewidth]{figures/multi_mod2.png}
%   \includegraphics[width=1\linewidth]{figures/multi_mod4.png}
%   \includegraphics[width=1\linewidth]{figures/multi_mod6.png}
%   \caption{Multi-Modal Cost Surfaces} 
%   \label{fig:multi_modal_heatmaps}
% \end{figure}

\subsection{Motion Planning Metrics} \label{tab:mp_metrics}

% \begin{center}
\begin{figure}[h]
\begin{tabularx}{1.0\textwidth} { 
  | >{\raggedright\arraybackslash}X 
  | >{\centering\arraybackslash}X
  | >{\centering\arraybackslash}X
  | >{\centering\arraybackslash}X 
  | >{\centering\arraybackslash}X 
  | >{\centering\arraybackslash}X 
  | >{\centering\arraybackslash}X 
  | >{\centering\arraybackslash}X 
  | >{\raggedleft\arraybackslash}X | }
 \hline
 Models & Avg. Jerk (mpsss) & Avg Lat Accel (radpss) & Avg. Progress (m) & Collision 0.5s & Collision 1.0s & Collision 1.5s & Collision 2.0s \\
 \hline
 \hline
  Human  & 4.14  & 1.93 & 13.05 & 0.00 & 0.00 & 0.00 & 0.00 \\
 \hline
 \hline
 Polynomial Frenet  & 3.03  & 3.89 & 13.27 & 0.00 & 0.00 & 0.01 & 0.02 \\
 \hline
 NAF  & 1.59  & 4.10 & 10.44 & 0.00 & 0.00 & 0.03 & 0.11 \\
 \hline
%  NAF + $\sigma$-VAE + IL   & 1.53  & 3.05 & 13.21 \\
%  \hline
 $\sigma$-VAE  & 3.47  & 3.20 & 13.29 & 0.00 & 0.00 & 0.00 & 0.1 \\
 \hline
 FlowPlan  & 1.53  & 3.05 & 13.21 & 0.00 & 0.00 & 0.00 & 0.1 \\
 \hline
\end{tabularx}

\vspace{0.2cm}

\caption{Motion Planning Metrics: We compare the performance comparison of different sampling techniques used for generating low cost control trajectories on various planning metrics. This table provides an intuition into how FlowPlan was able to generate plans with lower cost. Specifically, plans generated by FlowPlan has lower Avg Jerk, Avg. Lat Accel, higher Avg. Progress and minimal collisions. }
\end{figure}

% \end{center}

% \section{Algorithm Box}
% \begin{algorithm}[h]
% %  \footnotesize
% \SetAlgoLined
%  \SetKwInOut{Input}{Input}
%  \SetKwInOut{Output}{Output}
%  \Input{Something}

%  \Output{Something else}
%   Hello
  
%  \For{$i \leftarrow \ 1$ \KwTo $Iter$}{
%  Do something\\
% Then do something\\
% then again do something\\
% }
%  \caption{FlowPlan}
%  \label{algo:full}
% \end{algorithm}
% \section{Multi-modal Cost Surface}
% \section{Extract uniform dataset not just 95\% straights}
% \section{Output of our model}

\end{document}